\documentclass[10pt,twocolumn,letterpaper]{article}

\usepackage[pagenumbers]{cvpr} 
\usepackage{graphicx}%
\usepackage{multirow}%
\usepackage{amsmath,amssymb,amsfonts}%
\usepackage{amsthm}%
\usepackage{mathrsfs}%
\usepackage[title]{appendix}%
\usepackage{textcomp}%
\usepackage{manyfoot}%
\usepackage{caption}
\usepackage{diagbox}
\usepackage[export]{adjustbox}
\usepackage{listings}%
\usepackage{epsfig}
\usepackage{placeins}
\usepackage{makecell}
\usepackage{float}
\usepackage{booktabs}%
\usepackage{bm}
\usepackage{setspace}
\usepackage{makecell}
\usepackage{marvosym, ifsym} 

\definecolor{cvprblue}{rgb}{0.21,0.49,0.74}
\usepackage[pagebackref,breaklinks,colorlinks,allcolors=cvprblue]{hyperref}

\def\paperID{1238} 
\def\confName{CVPR}
\def\confYear{2025}

\title{Panorama Generation From NFoV Image Done Right}

\author{Dian Zheng$^{1,*}$
\quad
Cheng Zhang$^{2}$
\quad
Xiao-Ming Wu$^{1}$\\
\quad
Cao Li\textsuperscript{3,\Letter}
\quad
Chengfei Lv$^{3}$
\quad
Jian-Fang Hu$^{1}$
\quad
Wei-Shi Zheng\textsuperscript{1,4,\Letter}\\\\
$^{1}$Sun Yat-sen University
\quad
$^{2}$Monash University
\quad
$^{3}$Alibaba Group\\
\quad
$^{4}$Key Laboratory of Machine Intelligence and Advanced Computing, Ministry of Education, China\\
\tt\normalsize\color{Magenta}\url{https://isee-laboratory.github.io/PanoDecouple/} \\\\
}

\begin{document}
\maketitle
\begin{abstract}
Generating 360-degree panoramas from narrow field of view (NFoV) image is a promising computer vision task for Virtual Reality (VR) applications. Existing methods mostly assess the generated panoramas with InceptionNet or CLIP based metrics, which tend to perceive the image quality and is \textbf{not suitable for evaluating the distortion}. In this work, we first propose a distortion-specific CLIP, named Distort-CLIP to accurately evaluate the panorama distortion and discover the \textbf{``visual cheating''} phenomenon in previous works (\ie, tending to improve the visual results by sacrificing distortion accuracy). This phenomenon arises because prior methods employ a single network to learn the distinct panorama distortion and content completion at once, 
which leads the model to prioritize optimizing the latter. To address the phenomenon, we propose \textbf{PanoDecouple}, a decoupled diffusion model framework, which decouples the panorama generation into distortion guidance and content completion, aiming to generate panoramas with both accurate distortion and visual appeal. Specifically, we design a DistortNet for distortion guidance by imposing panorama-specific distortion prior and a modified condition registration mechanism; and a ContentNet for content completion by imposing perspective image information. Additionally, a distortion correction loss function with Distort-CLIP is introduced to constrain the distortion explicitly. The extensive experiments validate that PanoDecouple surpasses existing methods both in distortion and visual metrics. 

\end{abstract}
\makeatletter
\renewcommand*{\@makefnmark}{}
\footnotetext{
  $^*$ The work is done when Dian Zheng is an intern at Alibaba Group. \\
  \hspace*{1.8em}\textsuperscript{\Letter} corresponding authors. \hspace{5pt}\href{https://github.com/iSEE-Laboratory/PanoDecouple}{Code} is available
}
\begin{figure}
    \centering
    \includegraphics[width=0.98\linewidth]{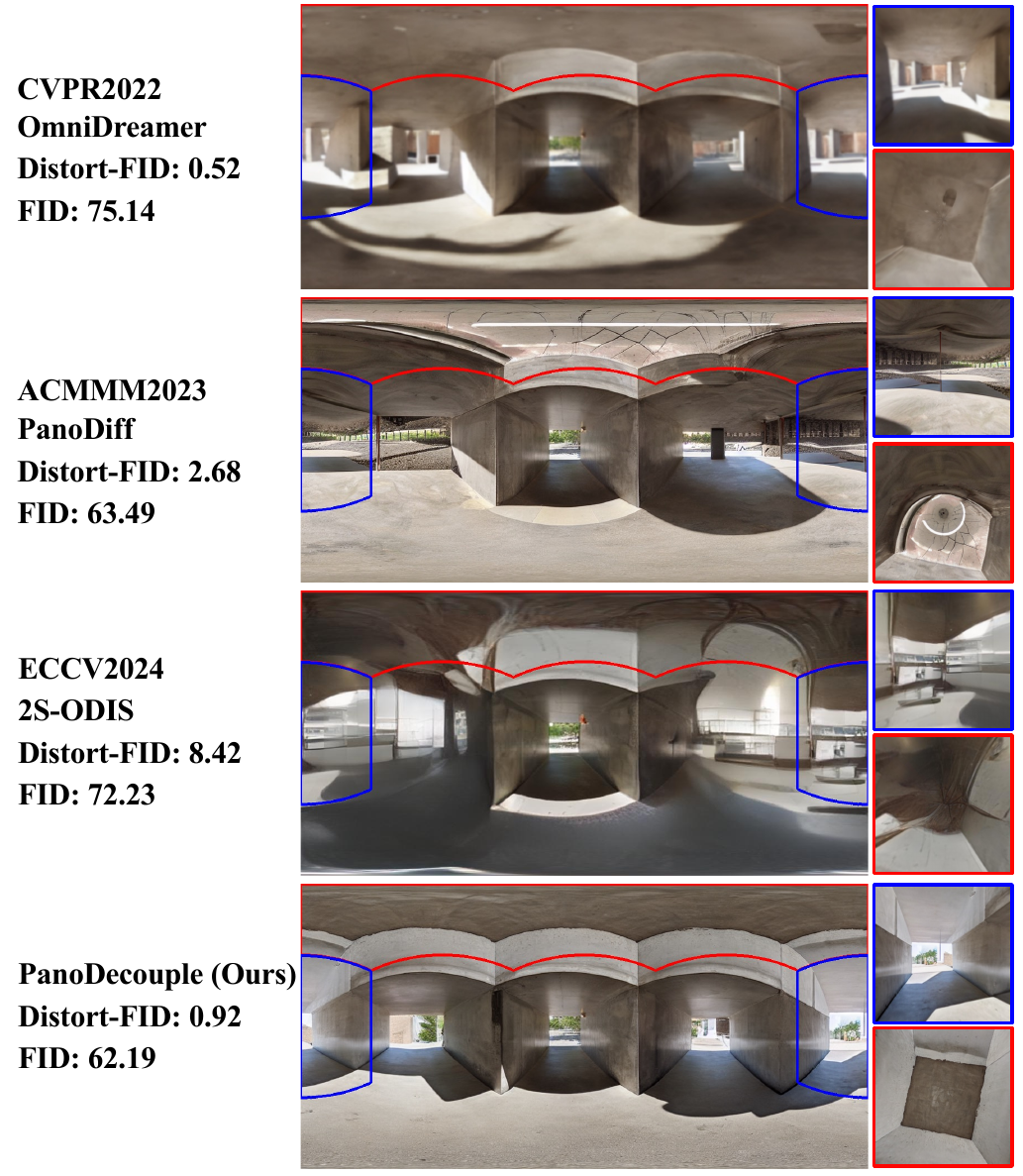}
    \vspace{-2mm}
    \caption{The image quality and distortion accuracy of existing methods and ours by FID and Distort-FID (ours) respectively. We project two regions in panorama (signed in corresponding color) into perspective image to show the distortion accuracy of existing methods (\ie, no distortion and natural layout in perspective image means good results). Recent methods improve the image quality while significantly ruining the distortion. We named it ``visual cheating'' phenomenon. Zoom in for best view.}
    \vspace{-6mm}
    \label{fig:motivation}
\end{figure}

\section{Introduction}
\label{sec:intro}
360-degree panorama generation from narrow field of view (NFoV) image aims to outpaint the partial panorama while maintaining accurate distortion and ensuring consistency in content, style with the NFoV image simultaneously, which has a wide range of applications in Virtual Reality (VR) and 3D scene generation~\cite{360gs, zhou2025dreamscene360}.

Existing methods mainly utilize InceptionNet~\cite{inception} or CLIP~\cite{clip} based evaluation metrics (\eg, FID, IS, \etc) to validate the generation performance. However, these models tend to perceive the image quality while ignoring the distortion as the feature similarity between panoramas with different contents is lower than panorama and perspective image with the same content (The detail is in Sec~\ref{sec:clip}). To address this, we propose a distortion-perception CLIP, named Distort-CLIP, which is tuned by different distortion types of data in contrastive learning mechanism. We substitute the InceptionNet used in FID with our Distort-CLIP and compute the Distort-FID of existing methods, observing the ``visual cheating'' phenomenon as shown in Fig~\ref{fig:motivation}. Surprisingly, OmniDreamer~\cite{omnidreamer} in 2022 achieves the best distortion accuracy while subsequent methods, misled by the existing evaluation metrics, blindly improve the image quality while gradually degrading distortion accuracy. We assume that this phenomenon is caused by learning distortion and content completion in one model and the model tends to prioritize optimizing the latter.

In this work, we propose a decoupled panorama generation pipeline, termed PanoDecouple, aiming to generate panoramas with accurate distortion and visual appeal at once. Specifically, we decouple the task of panorama generation into distortion guidance and content completion. For distortion guidance, we first introduce a general panorama distortion representation, distortion map into the network as explicit distortion guidance, then we modify the condition registration mechanism of ControlNet from first block only to all the blocks to make it suitable for position-encoding-like condition input (\ie, distortion map). As for content completion, we basically follow the architecture of mask-based outpainting methods~\cite{zhang2023adding} but substitute the text condition with perspective image information to better align the content of the generated panorama with the NFoV image. The two branches will be fused into the pre-trained, frozen U-Net, leveraging the powerful prior knowledge of pre-trained data and handling the information fusion simultaneously. What's more, we design a distortion correction loss, utilizing the distortion prior in Distort-CLIP to further constrain the distortion.

By introducing the decoupled pipeline, we achieve the best image quality and second-best distortion accuracy in SUN360 and SOTA performance in Laval Indoor in zero-shot manner. Notably, we only use 3K training data, which is 15 times less than the existing methods, but achieved surprising generalization ability, highlighting the importance of decoupling the task of panorama generation. To further validate the universality of our PanoDecouple, we propose two practical applications (\ie, text editing and text-to-image panorama generation) and show appealing results.

In summary, our main contributions are as follows:

$(1)$ We identify the limitation in current evaluation metrics for assessing panorama distortion and introduce a distortion-aware model, Distort-CLIP and corresponding evaluation metrics Distort-FID. Based on Distort-CLIP, we further observe the ``visual cheating'' phenomenon in existing methods and trace its cause to learn panorama distortion and content completion in a single model.

$(2)$ A decoupled panorama generation pipeline, PanoDecouple is proposed. PanoDecouple introduces a novel distortion guidance branch, DistortNet into the network with an accurate distortion prior and a corresponding condition registration mechanism. 

$(3)$ Our PanoDecouple surpasses prior methods on two benchmarks with only 3K training data. PanoDecouple can also be applied to text editing and text-to-image panorama generation for free with appealing results.

\begin{figure*}
    \centering
    \includegraphics[width=0.98\linewidth]{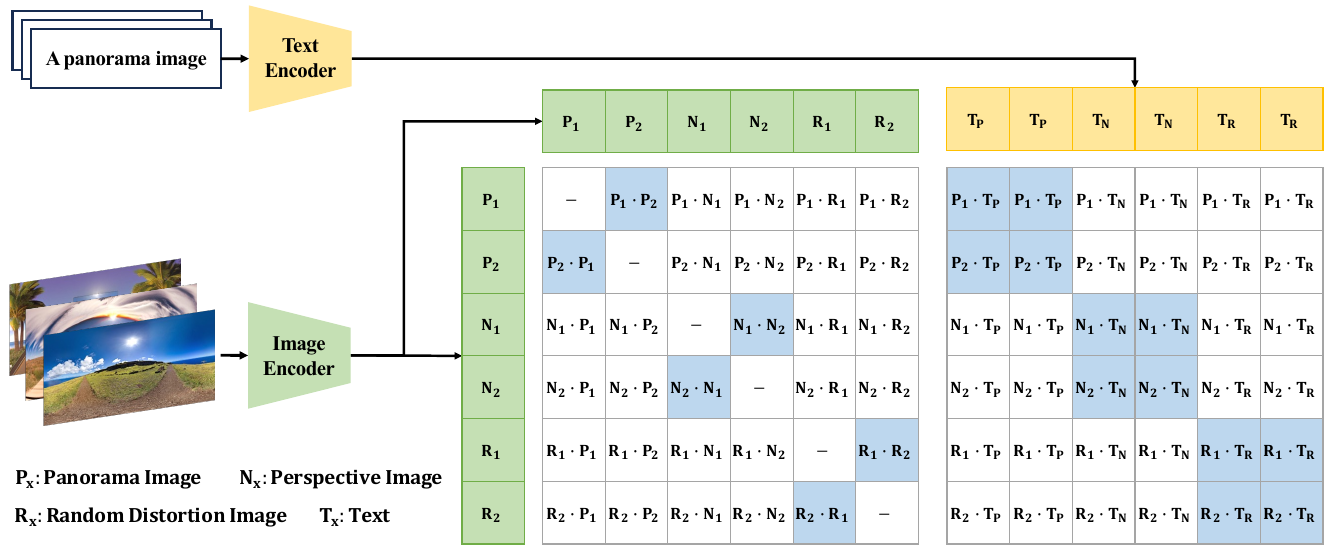}
    \vspace{-2mm}
    \caption{The training pipeline of our Distort-CLIP. The image features of three distortion types will do cosine similarity with themselves, and text features of three distortion types respectively. ``-'' means that the corresponding elements will not participate in the computation because it is meaningless. The boxes in \textcolor{blue}{blue} mean the similarity of corresponding elements is 1, otherwise 0. Zoom in for best view.}
    \label{fig:clip}
    \vspace{-4mm}
\end{figure*}

\vspace{-2mm}
\section{Related Work}
\label{sec:relate}
\vspace{-1mm}
\subsection{Image Outpainting}
\vspace{-1mm}
Image outpainting~\cite{gani, queryo, gardias2020enhanced, lu2021bridging, vqgan, chang2022maskgit, stable_diffusion, zhang2023adding} is a challenging computer vision task, and is an important step toward Artifcial General Intelligence~\cite{zheng2025diffuvolume, lin2023diversifying, lin2024human, zhou2021graph, zheng2024selective, lv2024spatialdreamer, liu2023generating, xu2024dexterous, wu2024economic, wu2023estimator}, which requires expanding a partial image with content and style consistency. Unlike inpainting, outpainting struggles to utilize information from pixels near the unknown area. SRN~\cite{gani} first leverages GAN~\cite{gan} for outpainting with the guidance of incomplete semantic information; QueryOTR~\cite{queryo} splits the image into patches, considering it as queries and utilize vision transformer~\cite{vaswani2017attention, vit} to outpaint the unknown patches assisted by queries. With the development of VQ-VAE~\cite{vae, vqvae, vqgan}, diffusion model~\cite{ddpm, ddim, stable_diffusion, dit, sit, sd3} and ControlNet~\cite{zhang2023adding, controlplus}, mask-based diffusion outpainting methods apply extra conditions into ControlNet and achieve great performance by leveraging the powerful prior knowledge of latent diffusion.

Although the methods above can achieve great results in content completion, they are not suitable for panorama generation because of the characteristics of panorama generation, which consists of distortion guidance and content completion. In this work, We design a decoupled diffusion model, consisting of a DistortNet and a ContentNet to address the two characteristics respectively.

\subsection{Panorama Generation}
\vspace{-1mm}
Panorama generation contains two settings: text-to-image generation and image outpainting. Text-to-image generation~\cite{chen2022text2light, Tang2023mvdiffusion, bar2023multidiffusion, li2023panogen, lee2023syncdiffusion, zhang2023diffcollage, stitch, panfusion} free the constrain of image input and fully utilize the prior knowledge of pre-trained latent diffusion model to generate fancy panorama. Text2Light~\cite{chen2022text2light} proposes distortion map, a general panorama distortion representation to constrain the generated distortion by considering it as the prompt tokens in the transformer. PanFusion~\cite{panfusion} sets the distortion map as position encoding and utilizes its information in an attention mechanism. As for panoramic image outpainting~\cite{omnidreamer, panodiff, wu2023panodiffusion, aog, odis}, in this setting, there's a greater emphasis on ensuring the consistency of the generated panorama with the content and style of the given NFoV image, while also maintaining accurate panoramic distortion. However, existing methods mainly consider the image quality of the generated panorama, while ignoring the distortion and causing ``visual cheating'' phenomenon as shown in Fig~\ref{fig:motivation}.

In this work, we adopt the distortion map followed~\cite{chen2022text2light, panfusion} to constrain the distortion accuracy. However, we do not leverage it in attention mechanism, but modify the original ControlNet architecture and make it suitable for position-encoding-like condition input (\ie, distortion map).
\vspace{-2mm}
\section{Distort-CLIP}
\vspace{-1mm}
\label{sec:clip}
In this section, we first show the motivation for our distortion analysis methods, then we generate data needed for evaluation and use the generated data to show the limited ability of existing metrics for evaluating distortion. Finally, we propose Distort-CLIP to focus on assessing the panorama distortion.

\subsection{Existing Metrics Evaluation}
\noindent\textbf{Motivation.} The best choice for evaluating the models' distortion assessment ability (used in existing metrics) is comparing the similarity of image pairs of ``different contents, same distortion'' and ``same contents, different distortion''. A good model will exhibit high feature similarity for image pairs with different content but the same distortion, and low similarity otherwise.

\noindent\textbf{Data Generation.} We aim to generate no-distortion (perspective) images that are consistent with the panoramic content to validate the model's capability systematically. Specifically, we first extract the central region of panoramas data (\ie, SUN360~\cite{sun360}) and project it into perspective images followed~\cite{panfusion}, then we apply sd-outpainting~\cite{zhang2023adding} model to outpaint the perspective images with the same text as panoramas but an extra prefix ``A perspective image of''.

\begin{table}[t]
    \centering
    \caption{Comparison of our Distort-CLIP with other models used in evaluation metric. We show the feature similarity (range from -1 to 1) between different pairs (\ie, different distortion, same content; same distortion, different content; panorama and different distortion texts). The best results are in \textbf{bold}.}
    \vspace{-2mm}
    \resizebox{0.95\columnwidth}{!}{
        \begin{tabular}{l|c|c|c}
        \toprule[0.12em]
        \textbf{Method} &
        \textbf{InceptionNet} & \textbf{CLIP} & \textbf{Distort-CLIP} \\
        \midrule[0.12em]
        Pano-Pers $\downarrow$ & 0.490 & 0.752 & \textbf{0.001} \\
        Pano$_{\text{train}}$-Pano$_{\text{test}}$ $\uparrow$ & 0.229 & 0.693 & \textbf{0.993}\\
        Pano$_{\text{train}}$-Text$_{\text{pano}}$ $\uparrow$ & - & 0.271 & \textbf{0.995} \\
        Pano$_{\text{train}}$-Text$_{\text{pers}}$ $\downarrow$ & - & 0.269 & \textbf{0.001} \\
        \bottomrule[0.1em]
    \end{tabular}{}
    }
    \label{tab:clip}
    \vspace{-4mm}
\end{table}

\noindent\textbf{Results.} We show the feature similarity between panoramas and perspective images with same content, panorama and panorama with different content in Table~\ref{tab:clip}. The results show that the InceptionNet (\ie, used in FID, IS) and CLIP (\ie, used in CLIP-FID, CLIP-Score) \textbf{tend to perceive image content information}, with a fragile ability to perceive distortions, especially in the case of InceptionNet. So we claim that existing evaluation metrics used in panorama generation are not suitable for distortion assessment and a more robust model is needed.

\subsection{Distort-CLIP Tuning} Based on the result in Table~\ref{tab:clip}, CLIP owns a certain ability to perceive the distortion and we think that it benefits from the contrastive learning mechanism. So we fine-tune the CLIP with panoramas and our generated images in contrastive learning mechanism. Here we add an additional random distortion types images that are consistent with the panoramic content to improve the robustness of the model. For the details of generating random distortion images, please refer to \textit{Appendix A}. Since this model is used for metric evaluation and subsequent distortion correction loss, it doesn't require generalization. Therefore, we directly fine-tune the CLIP network as shown in Fig.~\ref{fig:clip}. We show the details below.

\noindent\textbf{Image Encoder.} The image encoder is used to calculate the evaluation metric, which aims to distinguish the panoramic distortion in the image. So we input the three types of images (\ie, normal, random distortion, panoramic images) into the network at the same time and the training objective is to optimize the cosine similarity between the image pairs with same distortion type while enlarging the difference with others. The contrastive loss is defined as follows:
\vspace{-2mm}
\begin{equation} 
\mathcal{L}_{\text{ie}}(\bm{x}, \bm{y}) = 
\frac{1}{N\left(N-1\right)} \sum_{i=1}^{N} \sum_{j=1,j\neq i}^{N} \left(\bm{x}_i^\mathsf{T} \bm{y}_j - l_{ij}\right)^2,
\end{equation}
where $\bm{x}_i$, $\bm{y}_j$ are the normalized image vectors, $l_{ij}$ means the label of this index (\ie, 1 if $\bm{x}_i$ and $\bm{y}_j$ are the same type and 0 otherwise). We discard the self-similarity calculation (diagonal), as it holds no practical significance.

\begin{figure*}
    \centering
    \includegraphics[width=0.92\linewidth]{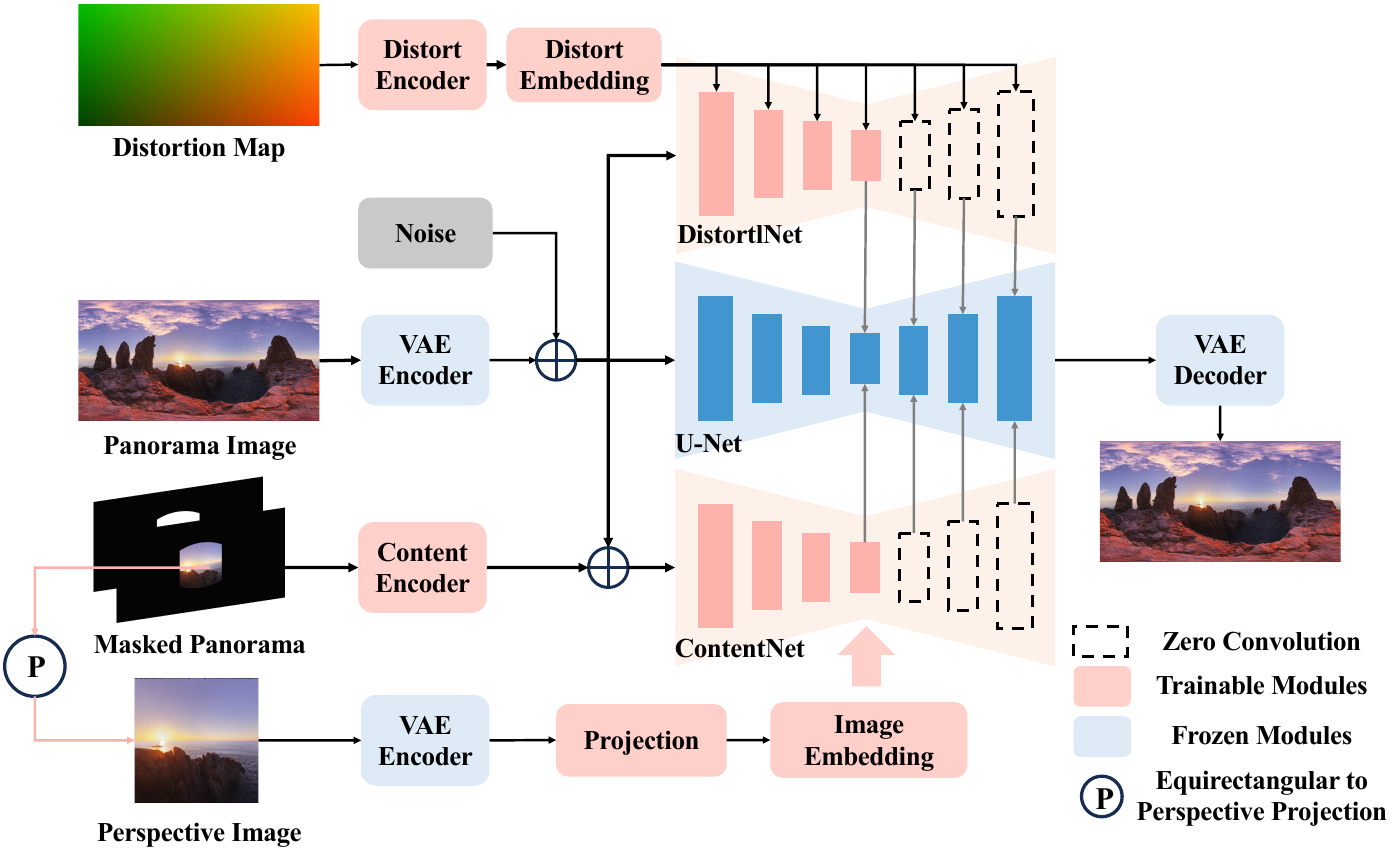}
    \vspace{-2mm}
    \caption{The pipeline of the proposed PanoDecouple, a decoupled diffusion model. The DistortNet focuses on distortion guidance via the proposed distortion map. To make full use of the position-encoding-like distortion map, we modify the condition registration mechanism of ControlNet from the first block only to all the blocks. The ContentNet is devoted to content completion by imposing partial panorama image input and perspective information. The U-Net remains frozen, coordinating the information fusion between content completion and distortion guidance branches, while fully leveraging its powerful pre-trained knowledge. Note that we omit the text input of the DistortNet and U-Net for simplification while the one for ContentNet is replaced by perspective image embedding.}
    \label{fig:pipeline}
    \vspace{-4mm}
\end{figure*}

\noindent\textbf{Text Encoder.} The text encoder is utilized to compute the distortion correction loss function (see Sec~\ref{sec:dloss} for details). This process requires merely distinguishing distortions rather than providing a detailed description of the image. Therefore, we only use three text prompts as input (\ie, ``A panorama image'', ``A perspective image'' and ``A random distortion image'') to fine-tune the text encoder. The contrastive loss is defined as follows:
\vspace{-2mm}
\begin{equation} 
\mathcal{L}_{\text{te}}(\bm{x}, \bm{z}) = 
\frac{1}{N\times N} \sum_{i=1}^{N} \sum_{j=1}^{N} \left(\bm{x}_i^\mathsf{T} \bm{z}_j - l_{ij}\right)^2,
\vspace{-2mm}
\end{equation}
where $\bm{z}_j$ is the normalized text vectors, the $l_{i,j}$ in $\mathcal{L}_{\text{te}}$ is the same as the one in $\mathcal{L}_{\text{ie}}$ as the distortion type of the image and text corresponds to each other. Additionally, recognizing the distinct roles of the two encoders, we cut off the gradient flow to the image encoder while tuning the text encoder. This strategy enables each encoder to achieve its full potential. The whole loss function is shown as follows:
\begin{equation}
    \mathcal{L} = \mathcal{L}_{\text{te}} + \mathcal{L}_{\text{ie}}.
\end{equation}

\noindent\textbf{Results.} As shown in Table~\ref{tab:clip}, after fine-tuning the CLIP, we significantly improve the ability to distinguish the distortion both in image-image and image-text manners. Note that the test set of SUN360 (Pano$_\text{test}$) is not covered within the training data range, yet Distort-CLIP can still accurately determine that their distortion types are the same, showing the robustness of our method.
We further show the distortion correction ability of our Distort-CLIP in the experiment~\ref{sec:ablate}.

\section{PanoDecouple}
To avoid the ``visual cheating'' phenomenon in previous work, we design PanoDecouple, decoupling the task of panorama generation into distortion guidance (DistortNet) and content completion (ContentNet) as shown in Fig~\ref{fig:pipeline}. We further propose the distortion correction loss based on the proposed Distort-CLIP to constrain the distortion explicitly. 

\subsection{Preliminary}
\label{sec:pre}
\noindent\textbf{Latent Diffusion.}
Latent diffusion is first introduced by \cite{stable_diffusion} to address the high cost of operations in image space. This pipeline first transforms the image $x$ into latent map $z$ via an autoencoder $\mathcal{E}$ and then trains a U-Net network $\epsilon_\theta$ to learn the denoising ability. The training objective is defined as follows:
\vspace{-2mm}
\begin{equation}
\label{eq:rec}
\mathcal{L} = \mathbb{E}_{\mathcal{E}(x), t, \epsilon, c_t} \left[ \left\| \epsilon - \epsilon_\theta(z_t, t, c_t) \right\|_2^2 \right],
\vspace{-2mm}
\end{equation}
where $z_t$ means the noised latent map in the timestep $t$ and $c_t$ is the text condition. In the reverse process, $z_T$ is initialized by standard Gaussian noise and is fed into $\epsilon_\theta$ iteratively to obtain the clean latent map $z_0$. Finally, the decoder $\mathcal{D}$ is applied to decode the $z_0$ into image space.

\noindent\textbf{ControlNet-based Panorama Outpainting.} ControlNet~\cite{zhang2023adding} provides a way for diverse conditions control by adding a control branch, and the adding mechanism in mask-based panorama outpainting is as follows:
\vspace{-1mm}
\begin{equation}
out = \mathcal{F}_m(z) + \mathcal{Z}(\mathcal{F}_{cn}(z, c_t, c_p, \mathcal{M})),
\vspace{-2mm}
\end{equation}
where $\mathcal{F}_m$ is the main U-Net, $\mathcal{F}_{cn}$ is the ControlNet which initializes by the parameters of U-Net encoder part. $\mathcal{Z}$ is the zero convolution layer, which aims to avoid large modifications in the early stage. $z$, $c_t$, $c_p$ $\mathcal{M}$ are the latent feature, text, partial panorama and the outpainting region mask respectively. \textbf{Note that conditions $c_p$ and $\mathcal{M}$ are only added in the first block in the source code of ControlNet} and the others only have text condition input as follows:
\vspace{-2mm}
\begin{equation}
\label{eq:control}
\begin{aligned}
    &\mathcal{F}_{cn}(z, c_t, c_p, \mathcal{M}):
    \left\{
    \begin{aligned}
        &cn^0 = CN(z, c_t) + CE(c_p, \mathcal{M}),\\
        &cn^b = CN(cn^{b-1}, c_t), \quad b\in (1, B),
    \end{aligned}
    \right.
\end{aligned}
\end{equation}

where $CE$ and $CN$ are the content encoder and ControlNet similar to ours in Fig~\ref{fig:pipeline}, $B$ is the block number.

\subsection{DistortNet}
Panorama is a unit sphere in 3D space, which is also the resource of the distortion. Therefore, a reasonable choice for achieving distortion guidance is searching for an explicit distortion representation, which allows the model to perceive the relationship between corresponding points in 2D image space and 3D space. We introduce a suitable representation, distortion map below and then design a DistortNet to fully leverage the potential of the distortion map.

\noindent\textbf{Distortion Map Construction.} The distortion map $D$ is built followed~\cite{chen2022text2light,panfusion}, which is a 2D position signals for 3D scene representation $S$ (\ie, panorama is a unit sphere in 3D space). Given a panoramic image $I \in \mathbb{R}^{H \times W \times 3}$, the pixel relationship with sphere S is defined as follows:
\vspace{-1mm}
\begin{equation}
\begin{aligned}
&S(\theta, \phi, r) = I(i, j), \\
&\theta = (2i/H - 1)\pi, \\
&\phi = (2j/W - 1)\pi/2, \\
&r = 1,
\end{aligned}
\vspace{-2mm}
\end{equation}
where $i$, $j$ is the pixel in panorama, $\theta$ and $\phi$ represent the azimuth angle (range from -$\pi$ to $\pi$) and the elevation angle (range from -$\pi/2$ to $\pi/2$) of the sphere, respectively. The center of the sphere $S$ is the position of the camera. So the distortion map $D \in \mathbb{R}^{H \times W \times 2}$ is defined as follows:
\vspace{-1mm}
\begin{equation}
    D(i, j) = (\theta, \phi).
\vspace{-2mm}
\end{equation}
The formula defines the correspondence between a pixel on a panorama image and a point on a 3D sphere. What's more, since a panorama is a sphere in 3D space, its left and right edges can seamlessly align in a 2D representation. However, the current distortion map $D$ does not exhibit this property (\ie, -$\pi$ and $\pi$ in the leftmost and rightmost edge). Therefore, we introduce first-order Taylor expansion positional encoding~\cite{nerf} $\gamma(\cdot)$ to make it continuous as follows:
\begin{equation}
\begin{aligned}
    &D(i, j) = (\gamma(\theta), \gamma(\phi)),\\
    &\gamma(\theta) = [sin(2^0\pi\theta), cos(2^0\pi\theta)],\\
    &\gamma(\phi) = [sin(2^0\pi\phi), cos(2^0\pi\phi)].
\end{aligned}
\end{equation}
In this way, the edges are continuous (sin(-$\pi$) and sin($\pi$), cos(-$\pi$) and cos($\pi$) are equal). The obtained $D \in \mathbb{R}^{H \times W \times 4}$ will be fed into DistortNet and we name it $c_d$ for uniform.

\noindent\textbf{Condition Registration Mechanism Modification.} As mentioned in Sec~\ref{sec:pre} Eq.~\ref{eq:control}, this operation benefits condition types like image, depth, pose but not position-encoding-like distortion map as commonly the position encoding should interact with model in each block (\eg, timestep $t$ in diffusion model~\cite{ddpm}, position embedding in ViT~\cite{vit}). 

In this work, we modify the condition registration mechanism from the first block only to all the blocks as follows:
\vspace{-1mm}
\begin{equation}
\begin{aligned}
    &\mathcal{F}_{dn}(z, c_t, c_d):
    \left\{
    \begin{aligned}
        &dn^b = DN(dn^{b-1}, c_t) + \mathcal{Z}(Proj^b(de)), \\
        &de = DE(c_d),
    \end{aligned}
    \right.
\end{aligned}
\end{equation}

where $DN$, $DE$ and $dn$ are the ``DistortNet'', ``Distort Encoder'' and ``Distort Embedding'' in Fig~\ref{fig:pipeline} respectively, $Proj^b$ is the 2D convolution to map the $de$ into dimension in $b^{th}$ block.
We further compare our modification with attention and original ControlNet mechanism in experiment.

\begin{table*}[t]
    \centering
    \caption{Comparison with SOTA methods. $\dagger$ means re-implementing in our setting for fair comparison. Note that the bottom region of Laval is entirely black edges and we crop 20\% of it when testing image quality and undo it when testing distortion as it requires full image. \textcolor{gray}{($\cdot$)} means the crop setting of PanoDiff (crop 20\% up and bottom region). The best, second-best results are in \textbf{bold}, \underline{underline}.}
    \vspace{-2mm}
    \resizebox{2\columnwidth}{!}{
        \begin{tabular}{ll|c|c|cccc|cccc}
        \toprule[0.15em]
        \multicolumn{2}{l|}{\multirow{2}{*}{\textbf{Method}}} & \multicolumn{1}{c|}{\multirow{2}{*}{\textbf{Year}}} & 
        \multicolumn{1}{c|}{\multirow{1}{*}{\textbf{Training}}} & \multicolumn{4}{c|}{\textbf{SUN360}} & 
        \multicolumn{4}{c}{\textbf{Laval Indoor}} \\
        \multicolumn{2}{c|} {} & {} & \textbf{samples} & FID $\downarrow$ & CLIP-FID $\downarrow$ & Distort-FID $\downarrow$ & IS $\uparrow$ & FID $\downarrow$ & CLIP-FID $\downarrow$ & Distort-FID $\downarrow$ & IS $\uparrow$  \\
        \midrule[0.15em]
        \multicolumn{2}{l|}{OmniDreamer} & 2022 & 50K & 75.14 & 11.13 & \textbf{0.52} & 4.58 & 133.15 \textcolor{gray}{(118.8)} & 22.47 \textcolor{gray}{(11.21)} & \underline{2.95} & 2.82 \textcolor{gray}{(3.04)}\\
        \multicolumn{2}{l|}{ImmerseGAN} & 2022 & 250K & - & 9.26 & - & - & - & - \textcolor{gray}{(12.69)} & - & - \\
        \multicolumn{2}{l|}{PanoDiff $\dagger$} & 2023 & 3K & \underline{63.49} & \underline{7.04} & 2.68 & \underline{6.51} & \underline{80.63} \textcolor{gray}{(\underline{77.25})} & 10.26 \textcolor{gray}{(\underline{7.77})} & 13.53 & \underline{2.90} \textcolor{gray}{(\underline{3.17})}\\
        \multicolumn{2}{l|}{AOG-Net $\dagger$} & 2024 & 3K & 74.07 & 8.87 & 4.52 & 6.32 & 93.09 \textcolor{gray}{(89.38)} & 12.04 \textcolor{gray}{(9.25)} & 11.59 & 2.65 \textcolor{gray}{(2.89)}\\
        \multicolumn{2}{l|}{2S-ODIS $\dagger$} & 2024 & 3K+9K & 72.23 & 9.11 & 8.23 & 5.56 & 81.43 \textcolor{gray}{(80.67)} & \underline{9.45} \textcolor{gray}{(7.97)} & 10.21 & 2.54 \textcolor{gray}{(2.60)}\\
        \multicolumn{2}{l|}{\textbf{PanoDecouple (Ours)}} & - & 3K & \textbf{62.19} & \textbf{6.21} & \underline{0.92} & \textbf{6.93} & \textbf{74.45} \textcolor{gray}{(\textbf{72.51})} & \textbf{8.88} \textcolor{gray}{(\textbf{7.30})} & \textbf{2.85} & \textbf{2.95} \textcolor{gray}{(\textbf{3.28})}\\
        \bottomrule[0.1em]
    \end{tabular}{}
    }
    \vspace{-4mm}
    \label{tab:quan}
\end{table*}

\subsection{Decoupled Diffusion Model}
\vspace{-1mm}
\noindent\textbf{ContentNet.} The ContentNet basically follows the architecture of previous mask-based outpainting method~\cite{panodiff}. However, we substitute the text embedding with the proposed perspective image embedding to avoid the text-image inconsistent problem~\cite{chen2023pixart, moviegen, sd3} caused by BLIP-2~\cite{blip2}. Specifically, we use the Equirectangular-to-Perspective Projection as mentioned in \textit{Appendix} of~\cite{panfusion} to project the central region of panoramic image $c_p$ into the perspective image $c_n$ as shown in Fig~\ref{fig:pipeline}. Then the ``VAE Encoder'' and ``Projection'' networks are used to match the dimension with original text embedding as follows:
\vspace{-1mm}
\begin{equation}
    c_n' = Proj(VAE(c_n)).
\end{equation}

\noindent\textbf{Fusion in U-Net.} The functions of the main U-Net are leveraging the powerful prior knowledge and coordinating the information of distortion guidance and content completion branches simultaneously. The fusion pipeline of each block is defined as:
\vspace{-1mm}
\begin{equation}
out = \mathcal{F}_m(z) + \mathcal{Z}(\mathcal{F}_{cn}(z, c_n', c_p, \mathcal{M})) + \mathcal{Z}(\mathcal{F}_{dn}(z, c_t, c_d)),
\end{equation}

\subsection{Distortion Correction Loss}
\label{sec:dloss}
To generate more accurate distortion, we propose the distortion correction loss function via the text encoder of the Distort-CLIP to constrain the generation output explicitly. Specifically, we first transform the model output $\epsilon$ into clean latent code $z_0$ by leveraging the forward diffusion process:
\begin{equation}
    z_0 = \frac{1}{\sqrt{\overline{\alpha}_t}}(z_t - \sqrt{1-\overline{\alpha}_t}\epsilon),
\end{equation}
where $z_t$ is the noised latent code in timestep t and $\alpha$ is the noise coefficient. Then we use the VAE decoder $\mathcal{D}$ to decode $z_0$ into image space $x$ as follows:
\begin{equation}
    x = \mathcal{D}(z_0).
\end{equation}
After obtaining the generated panoramic image $x$, we compute its cosine similarity with the text of three types of distortion $\mathrm{T_P}$, $\mathrm{T_N}$ and $\mathrm{T_R}$, which is similar to Fig~\ref{fig:clip}. We set $\bm{x}$, $\bm{z_{\scriptscriptstyle P}}$, $\bm{z_{\scriptscriptstyle N}}$ and $\bm{z_{\scriptscriptstyle R}}$ as the normalized image and text feature vectors after Distort-CLIP encoders for simplification. The loss function is defined as follows:
\begin{equation}
    \mathcal{L}_{dist} = \bm{x}^\mathsf{T} \bm{z_{\scriptscriptstyle P}} - \bm{x}^\mathsf{T} \bm{z_{\scriptscriptstyle N}} - \bm{x}^\mathsf{T} \bm{z_{\scriptscriptstyle R}}.
\end{equation}
The full loss function for training our PanoDecouple is as follows:
\begin{equation}
    \mathcal{L} = \mathcal{L}_{rec} - \lambda\cdot\mathcal{L}_{dist},
\end{equation}
where $\mathcal{L}_{rec}$ is the same as Eq.~\ref{eq:rec} and $\lambda$ is the coefficient which we set at 0.05 in the experiment.

\section{Experiments}
\subsection{Experimental Setup}
\noindent\textbf{Datasets.} We follow PanoDiff~\cite{panodiff} to conduct experiments in SUN360~\cite{sun360} and Laval Indoor~\cite{laval} datasets. SUN360 comprises both indoor and outdoor scenes while Laval Indoor only has indoor scene. We use 3000/500 SUN360 data for training/testing and 289 Laval Indoor data for zero-shot generalization. \textbf{Note} that it is not the full number of SUN360 as the official link is unavailable and for the methods with a larger training number than ours, we directly test their pre-trained models on our test set to maintain fair comparison. As for the methods which does not open the pre-trained model, we re-implement them in our setting. The text prompts used in our methods are generated with BLIP-2~\cite{blip2}

\noindent\textbf{Evaluation Metrics.} As previous evaluation metrics exist problems in perceiving distortion. We utilize Fr\'{e}chet
 Inception Distance (FID)~\cite{fid}, CLIP-FID and Inception Score (IS)~\cite{is} to evaluate the panoramic image quality and use the proposed Distort-CLIP to compute the Distort-FID followed the calculation process of FID.
 
\noindent\textbf{Implementation Details.} PanoDecouple is trained based on the pre-trained weight of~\cite{zhang2023adding}, using one H20 GPU with PyTorch~\cite{paszke2019pytorch} environment. During training, We use the Adam~\cite{kingma2014adam} optimizer and L2 loss for 20 epochs with the initial learning rate 1e-5 and batch sizes 2. During inference, we use 50 timesteps for all the experiments. 

\begin{figure*}
    \centering
    \includegraphics[width=1\linewidth]{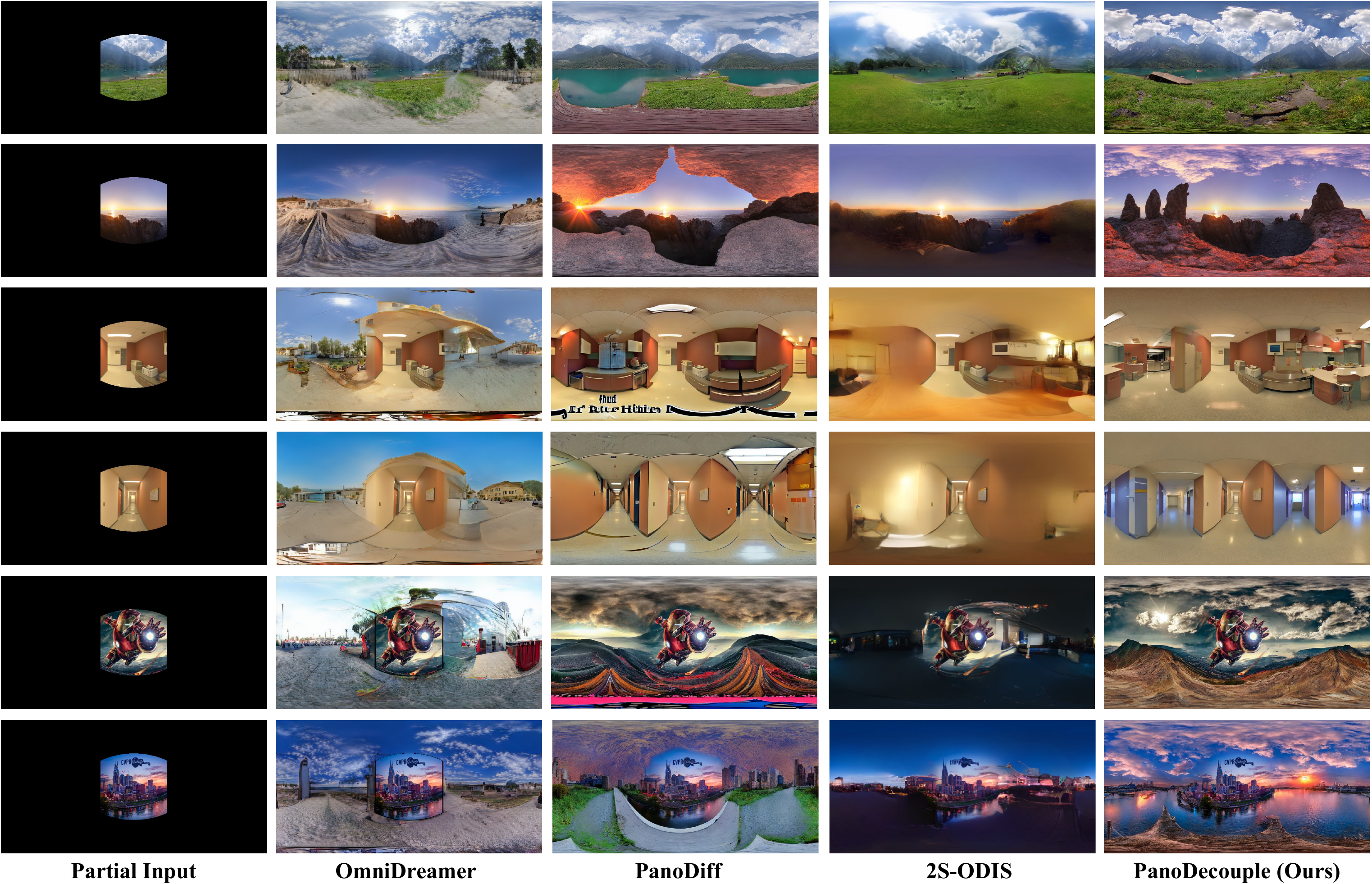}
    \vspace{-8mm}
    \caption{Qualitative comparison of panorama generation from NFoV image. We sequentially present the results on SUN360, Laval Indoor, and raw images (two images each). Zoom in for best view.}
    \vspace{-4mm}
    \label{fig:visual}
\end{figure*}

\subsection{Evaluation on the benchmarks}
\noindent\textbf{Quantitative Results.} The result is shown in Table~\ref{tab:quan}, in SUN360, the OmniDreamer~\cite{omnidreamer} achieves best distortion accuracy with limited image quality and subsequent methods, constrained by evaluation metrics, blindly improved image quality, gradually ruining distortion accuracy, which is the ``visual cheating'' phenomenon we pointed out. Our PanoDecouple significantly reduces the distortion inaccuracy in recent methods and achieves great image quality at the same time. In Laval Indoor, as the ceiling is hard to generate accurately, PanoDiff crops it while we maintain it as a harder setting. Notably, by using only 3K training data, we could achieve state-of-the-art image quality and great distortion accuracy, especially in Laval Indoor (zero-shot manner), showing that decoupling the panorama generation into distortion guidance and content completion truly benefits performance improvement and generalization ability.

\noindent\textbf{Qualitative Results.} The result is shown in Fig~\ref{fig:visual} (\ie, \textit{Appendix C} for perspective and more panorama results), both in SUN360, Laval Indoor and raw images, we achieve great image quality and distortion accuracy at once. Notably, we only use 3K training data, but showing great generalization ability even in raw website images. This phenomenon validates the significance of decoupling panorama generation into distortion guidance and content completion. 
\vspace{-1mm}
\subsection{Ablation Study}
\vspace{-1mm}
\label{sec:ablate}
We conduct ablation study on SUN360. We first evaluate the effectiveness of each component we proposed and validate the synergistic effect; then we explore the best condition registration mechanism for position-encoding-like distortion map; finally we prove the effectiveness of using general distortion representation, distortion map.

\begin{table}[t]
    \centering
    \setlength{\tabcolsep}{2.5pt}
    \caption{Quantitative comparison for the ablation study. SD, MD, PN, DLoss mean first-block condition registration in original ControlNet, all-block condition registration in our DistortNet, perspective image embedding and distortion correction loss respectively. The best, second-best results are in \textbf{bold} and \underline{underline}.}
    \vspace{-2mm}
    \resizebox{0.98\columnwidth}{!}{
        \begin{tabular}{ll|cccc|c|c|c}
        \toprule[0.15em]
        \multicolumn{2}{l|}{\textbf{Method}} & \textbf{SD} & \textbf{MD} & \textbf{PN} & \textbf{DLoss} &
        \textbf{CLIP-FID} $\downarrow$ & \textbf{Distort-FID} $\downarrow$ & \textbf{IS} $\uparrow$\\
        \midrule[0.15em]
        \multicolumn{2}{l|}{\multirow{5}{*}{\textbf{Ours}}} & & &  &  & 7.04 & 2.68 & 6.51 \\ 
        & & \checkmark &  &   &  & 6.66 & 1.63 & 6.16 \\
        & &  & \checkmark &   &  & \underline{6.38} & 1.05 & 6.35 \\
        & &  & \checkmark & \checkmark &   & 6.45 & \underline{0.98} & \textbf{6.95} \\
        & &  & \checkmark & \checkmark & \checkmark  & \textbf{6.21} & \textbf{0.92} & \underline{6.93}\\
        \bottomrule[0.1em]
    \end{tabular}{}
    }
    \label{tab:ablate}
    \vspace{-6mm}
\end{table}

\begin{figure*}
    \centering
    \includegraphics[width=0.95\linewidth]{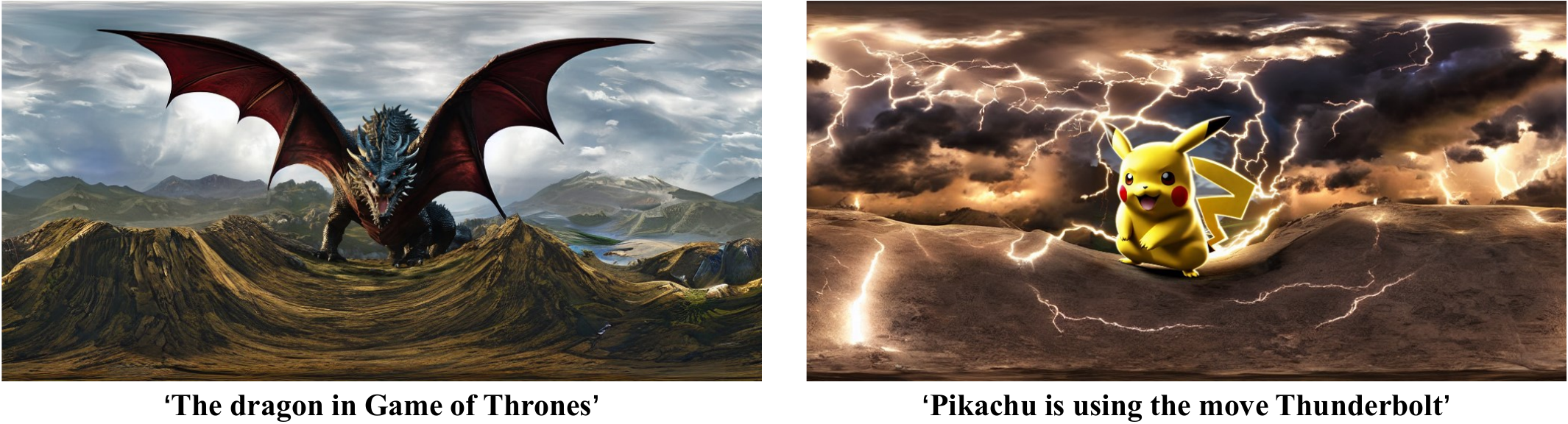}
    \vspace{-3mm}
    \caption{The quantitative results of text-panorama generation. Zoom in for best view.}
    \vspace{-6mm}
    \label{fig:edit}
\end{figure*}

\noindent\textbf{The Synergistic Effect Analysis.} We evaluate the effectiveness of the proposed DistortNet, ContentNet and distortion correction loss in Table~\ref{tab:ablate}. Note that without any modules we proposed is an image-outpainting diffusion model (PanoDiff~\cite{panodiff}). By feeding the distortion map into extra ControlNet in first-block condition registration mechanism, we significantly improve the distortion accuracy while degrading the generation diversity (IS) as adding additional condition naturally affects the diversity of network generation. When modifying the mechanism into all-block registration (\ie, our proposed DistortNet), we further improve the distortion accuracy. It could be seen that benefiting from decoupling the network, we ease the task of content completion and the image quality is gradually improved. When replacing the text embedding in ContentNet with perspective image embedding, we largely improve the generation diversity, benefiting from consistent representation between panoramas and NFoV images. Finally, by introducing the distortion correction loss into the model, not only the distortion accuracy is refined, but also the decoupling function is enhanced and simplifies the task of content completion, improving the image quality. In summary, the design of the decoupling diffusion model achieves a synergistic effect in both image quality and distortion constraint.

\noindent\textbf{Condition Registration Mechanism.} Here we systematically analyze which condition registration mechanism is suitable for position-encoding-like distortion map. We design four model architectures for analysis, the detailed architecture is shown in \textit{Appendix B} and the results are shown in Table~\ref{tab:regist}, utilizing all-block condition registration mechanism works better than the attention and original ControlNet mechanism. We hope this observation will contribute to advancing future research on diverse condition injection in diffusion models.

\noindent\textbf{Distortion Condition Selection.} To validate the effectiveness of distortion map, we replace it with full-zero map, learnable map and distortion map without first-order Taylor extension as shown in Table~\ref{tab:regist}, the learnable map achieves certain improvements while degrading the image diversity as learning the distortion map complicates the task. The introduced distortion map achieves the best result as it offers the network prior knowledge about panorama distortion representation, while without Taylor extension, the leftmost and rightmost consistency is ruined.

\begin{table}[t]
    \centering
    \caption{Quantitative comparison for the ablation study about the condition registration mechanism and distortion condition selection. The best results are in \textbf{bold}.}
    \vspace{-2mm}
    \resizebox{0.95\columnwidth}{!}{
        \begin{tabular}{ll|c|c|c|c}
        \toprule[0.15em]
        \multicolumn{2}{l|}{\textbf{Method}} & \textbf{FID} $\downarrow$ & 
        \textbf{CLIP-FID} $\downarrow$ & \textbf{Distort-FID} $\downarrow$ & \textbf{IS} $\uparrow$\\
        \midrule[0.15em]
        \multicolumn{4}{l}{\textit{Condition Registration Mechanism}} \\
        \midrule[0.1em]
        \multicolumn{2}{l|}{attn-unet} & 65.34 & 6.97 & 2.52 & 6.06 \\ 
        \multicolumn{2}{l|}{add-contorlnet} & 62.95 & 6.54 & 1.59 & 6.83 \\
        \multicolumn{2}{l|}{attn-contorlnet} & 65.71 & 7.92 & 5.22 & 5.58 \\
        \multicolumn{2}{l|}{attn-distortnet} & 63.54 & 6.56 & 1.48 & 6.57 \\
        \multicolumn{2}{l|}{\textbf{add-distortnet}}& \textbf{62.19} & \textbf{6.21} & \textbf{0.92} & \textbf{6.93}  \\
        \midrule[0.1em]
        \multicolumn{4}{l}{\textit{Distortion Condition Selection}} \\
        \midrule[0.1em]
        \multicolumn{2}{l|}{zero map} & 63.96 & 6.57 & 2.40 & 6.47 \\
        \multicolumn{2}{l|}{learnable map} & 63.32 & 6.43 & 1.42 & 6.23 \\
        \multicolumn{2}{l|}{distortion w/o Taylor}  & 62.57 & 6.68 & 2.48 & 5.78 \\
        \multicolumn{2}{l|}{\textbf{distortion map}} & \textbf{62.19} & \textbf{6.21} & \textbf{0.92} & \textbf{6.93}  \\
        \bottomrule[0.1em]
    \end{tabular}{}
    }
    \label{tab:regist}
    \vspace{-6mm}
\end{table}

\subsection{Applications}
\vspace{-1mm}
\noindent\textbf{Text Editing.} With different text prompts and the same NFoV image, our PanoDecouple can generate panoramas with different scenes and styles as shown in \textit{Appendix}. 

\noindent\textbf{Random NFoV Image Input.} PanoDecouple could achieve it by few step tuning based on original 90-degree central NFoV models. We show the result in \textit{Appendix}.

\noindent\textbf{Text-to-Panorama Generation.} Our PanoDecouple can freely apply to another setting: text-to-panorama generation by first utilizing a pre-trained text-to-image diffusion model, SDXL~\cite{sdxl} and then feeding the generated image into our PanoDecouple. The result is shown in Fig~\ref{fig:edit} bottom line. We generate panoramas with visual appeal, accurate distortion and text consistency simultaneously.

\vspace{-2mm}
\section{Conclusion}
\vspace{-2mm}
In this work, we experimentally discover the limitation of existing evaluation metrics in perceiving panorama distortion and propose a distortion-specific Distort-CLIP to address it. Based on Distort-CLIP, we further observe the visual cheating phenomenon in previous works which tends to improve the image quality while sacrificing the distortion accuracy. We think that this is caused by learning panorama distortion and content completion in a single model at once and propose a decoupled panorama generation pipeline, PanoDecouple, to address it. PanoDecouple consists of a DistortNet by imposing panorama distortion prior and a corresponding condition registration mechanism; and a ContentNet equipped with natural image information. Extensive experiments show that PanoDecouple achieves accurate distortion and visual appeal at once.

\noindent\textbf{Acknowledgments.} This work was supported partially by the National Key Research and Development Program of China (2023YFA1008503), NSFC(92470202, U21A20471), Guangdong NSF Project (No. 2023B1515040025); and Alibaba Group through Alibaba Research Intern Program.

{
    \small
    \bibliographystyle{ieeenat_fullname}
    \bibliography{main}
}

\maketitlesupplementary
\renewcommand{\thetable}{S\arabic{table}}
\renewcommand{\thefigure}{S\arabic{figure}}
\renewcommand\thesection{\Alph{section}}

\begin{figure}[t]
    \centering
    \includegraphics[width=1\linewidth]{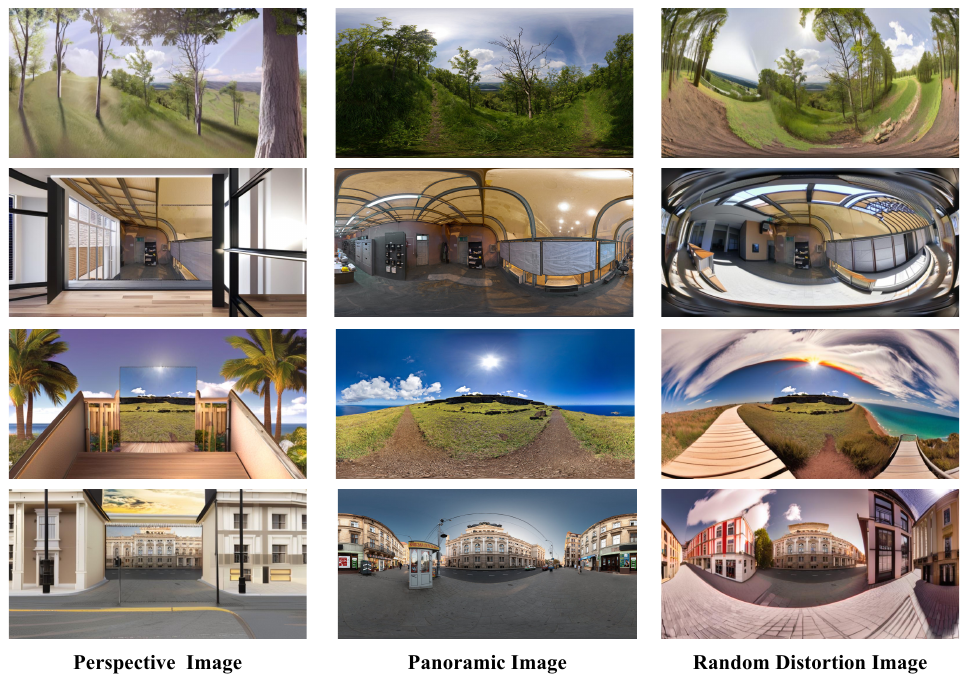}
    \vspace{-4mm}
    \caption{Visual results of panoramic images and corresponding perspective images. Zoom in for best view.}
    \vspace{-6mm}
    \label{fig:distortion}
\end{figure}

\section{Random Distortion Image Generation}
\vspace{-2mm}
\noindent\textbf{Motivation.} The random distortion image could accelerate the robustness of our Distort-CLIP to assess distortion. The challenge is how to generate it while sharing the same content with panorama. We observe that the training data of stable diffusion~\cite{sdxl} contains panorama image (with the height-width ratio 1:2). However, due to its relatively small proportion in training data, the distortion of the generated panoramic image is quite chaotic, which perfectly meets our requirements. 

\noindent\textbf{Method.} We apply a mask to the panoramic image, leaving only the central area (\ie, the same mask as shown in the main paper Fig. \textcolor{cvprblue}{3}). Then we use the sd-outpainting based on~\cite{sdxl} to outpaint the model with an extra prefix ``A panorama image of ''. In this way, we obtain random distortion images with the same content as panoramas.

\noindent\textbf{Visual results.} Here we also show the panoramic, generated perspective and random distortion images in Fig~\ref{fig:distortion}. It could be seen that we generate different distortion types images with similar content, ensuring the success of our Distort-CLIP.

\begin{figure}
    \centering
    \includegraphics[width=1\linewidth]{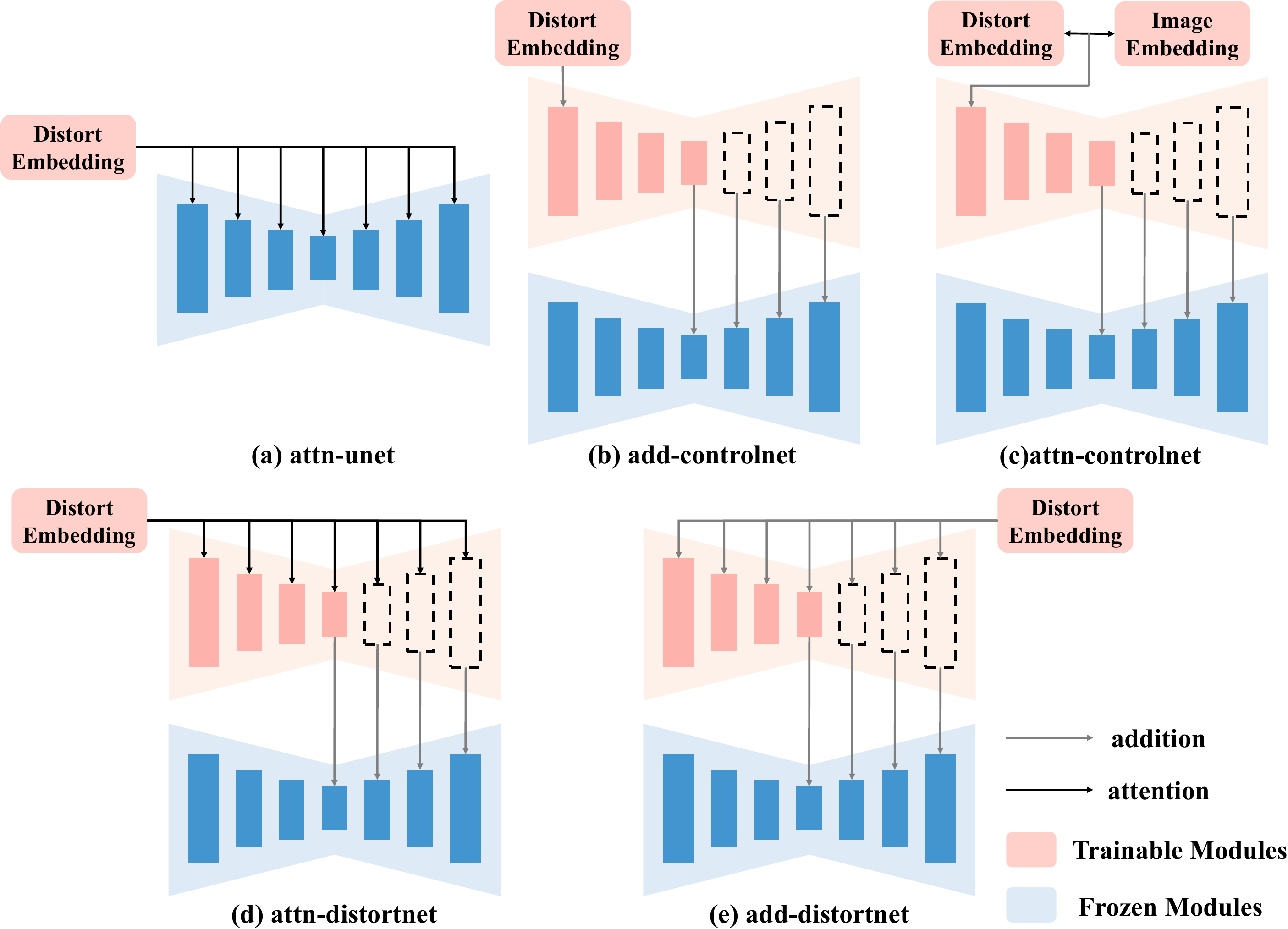}
    \caption{Various condition registration mechanisms. Note that both addition and attention own small learnable parameters (\eg, projection layers and zero convolution for addition; common attention modules for attention), we omit it in figure for simplification. Zoom in for best view.}
    \vspace{-4mm}
    \label{fig:arch}
\end{figure}

\section{Condition Registration Mechanism}
\vspace{-2mm}
We show the model architecture of different condition registration mechanisms in Fig~\ref{fig:arch}. We explore the widely used mechanism for condition registration and validate that the all-block registration with addition is the best choice for position-encoding-like conditions.

\section{More Applications}
\vspace{-2mm}
\noindent\textbf{Text Editing.} We show the text editing results below. Note that we specifically selected this NFoV image (with snow at the bottom), which is extremely challenging for text editing when the text is inconsistent with the snow. However, our method still generates panoramas consistent with the text.
\begin{figure}[h]
    \centering
    \includegraphics[width=1\linewidth]{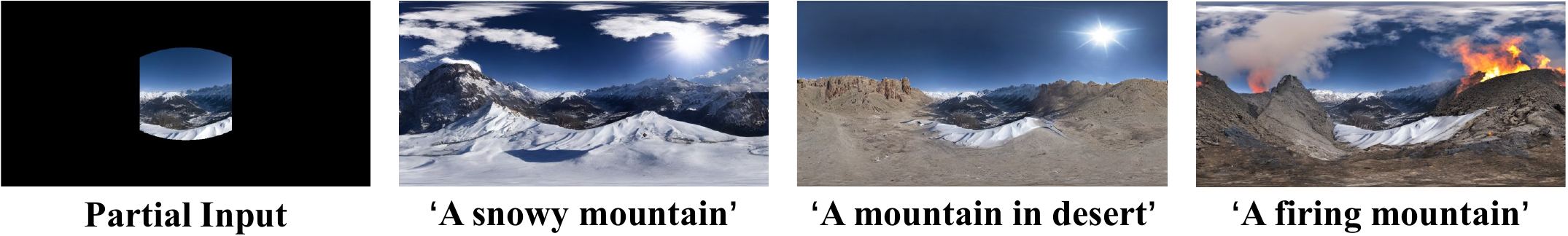}
    \vspace{-6mm}
    \caption{Visual results of text editing.}
    \label{fig:te}
\end{figure}

\noindent\textbf{Random NFoV.} We show results below, even the input image is extremely small and the position is tricky, we could generate panoramas with visual appealing.
\begin{figure}[h]
    \centering
    \includegraphics[width=0.95\linewidth]{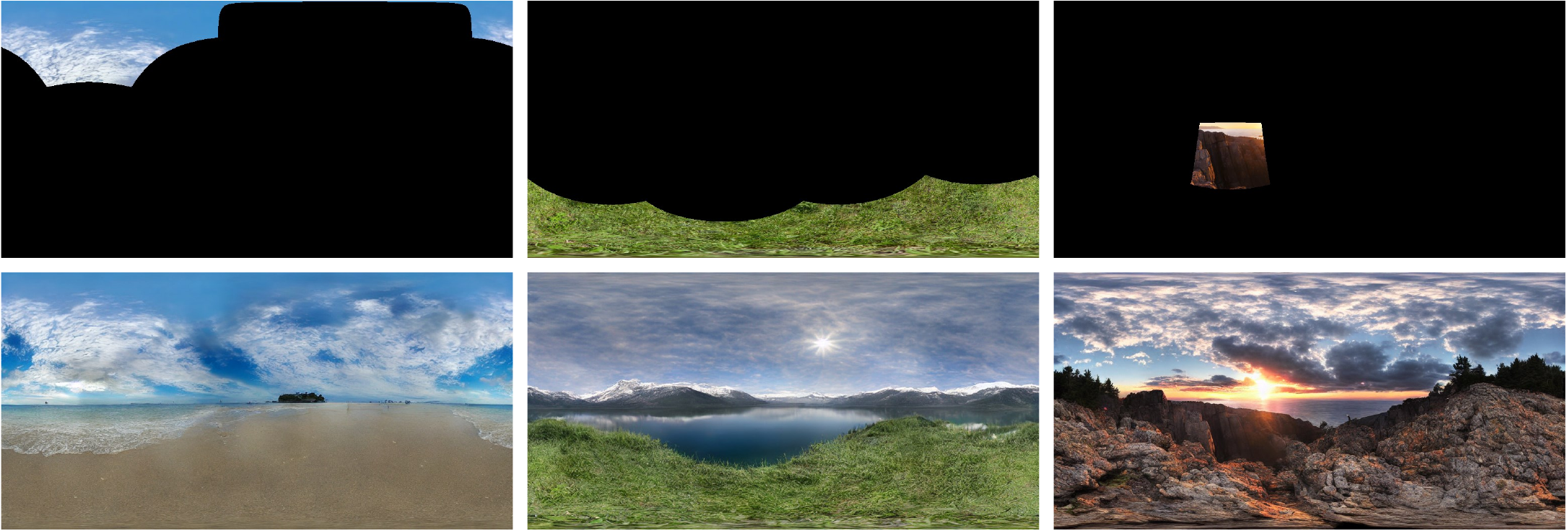}
    \vspace{-2mm}
    \caption{Visual results with random NFoV input.}
    \vspace{-2mm}
    \label{fig:fov}
\end{figure}

\section{More Visual Results}
\begin{figure*}
    \centering
    \includegraphics[width=0.95\linewidth]{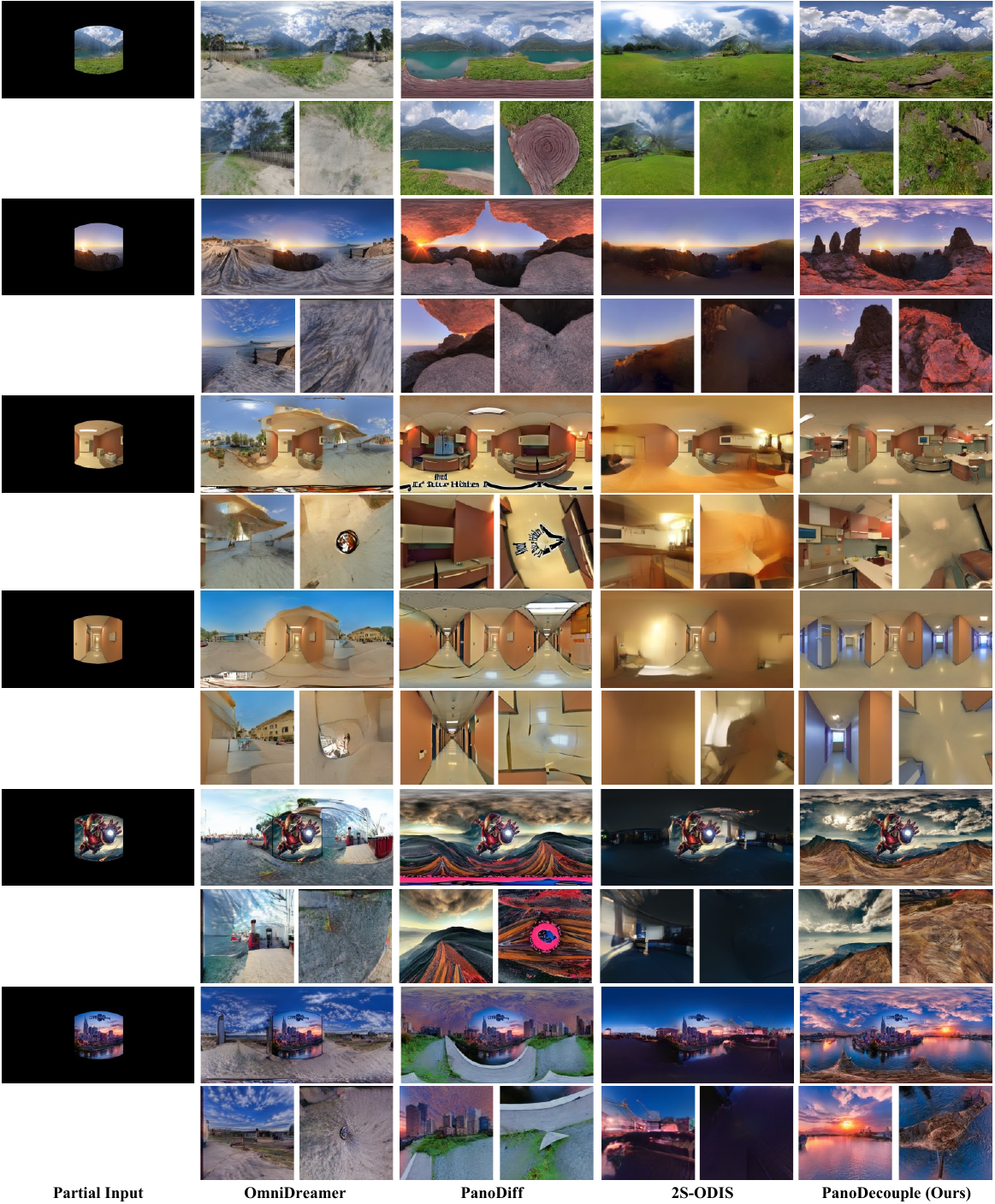}
    \caption{Visual results of panoramic images and corresponding perspective images.}
    \label{fig:sup_vis}
\end{figure*}

\begin{figure*}
    \centering
    \includegraphics[width=0.95\linewidth]{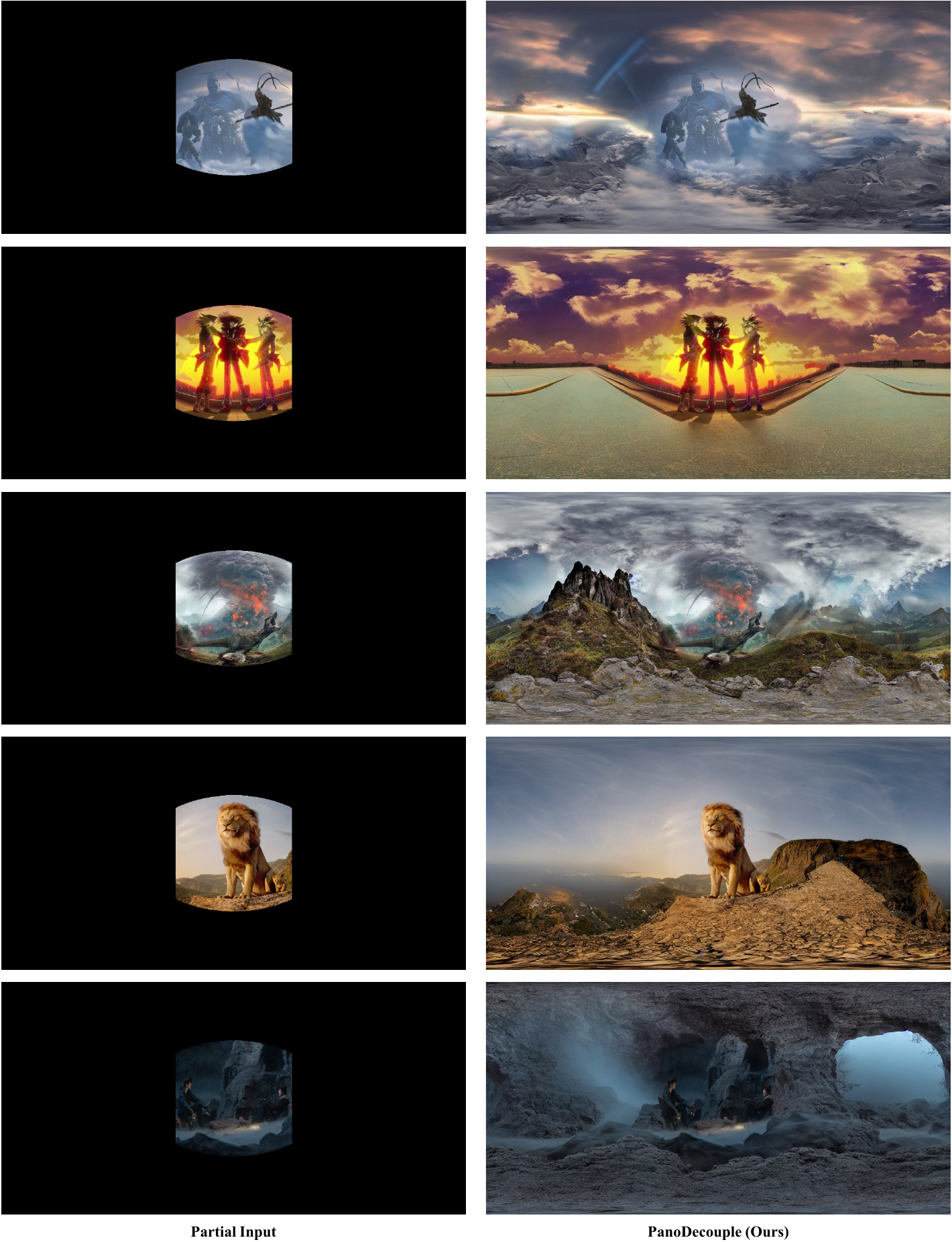}
    \caption{Visual results with raw image input. \textbf{Note that the images we use are for academic purposes only. If any copyright infringement occurs, we will promptly remove them}.}
    \label{fig:more}
\end{figure*}

In Fig~\ref{fig:sup_vis}, we show panoramic and perspective results of generated panoramas by different methods. Our PanoDecouple achieves great image quality and accurate distortion simultaneously.  We also show more raw image panorama outpainting results in Fig~\ref{fig:more}. Enjoy it!

\end{document}